\pdfoutput=1

\documentclass[11pt]{article}

\usepackage[preprint]{acl}

\usepackage{times}
\usepackage{latexsym}
\usepackage{float}
\usepackage{placeins}
\usepackage{stfloats}

\usepackage[]{xcolor} 
\usepackage{graphicx}
\usepackage{multirow}
\usepackage[breakable,skins]{tcolorbox}

\newtcolorbox{promptbox}[1][]{
    breakable,
    colback=gray!10,    
    colframe=gray!20,     
    title=#1,           
    fonttitle=\bfseries,
    boxrule=0.5mm,      
    arc=2mm,            
    outer arc=2mm,      
    coltitle=black,     
    enhanced,
}

\usepackage[T1]{fontenc}

\usepackage[utf8]{inputenc}

\usepackage{microtype}

\usepackage{inconsolata}

\usepackage{microtype}
\usepackage{graphicx}
\usepackage{subfigure}
\usepackage{booktabs} 

\usepackage{svg}
\usepackage{listings}
\usepackage{xcolor}
\usepackage{longtable}
\usepackage{calc}

\definecolor{codegreen}{rgb}{0,0.6,0}
\definecolor{codegray}{rgb}{0.5,0.5,0.5}
\definecolor{codepurple}{rgb}{0.58,0,0.82}
\definecolor{backcolour}{rgb}{0.95,0.95,0.92}

\lstdefinestyle{mystyle}{
    backgroundcolor=\color{backcolour},   
    commentstyle=\color{codegreen},
    keywordstyle=\color{magenta},
    numberstyle=\tiny\color{codegray},
    stringstyle=\color{codepurple},
    basicstyle=\ttfamily\footnotesize,
    breaklines=true,                 
    captionpos=b,                    
    keepspaces=true,                                    
    numbersep=5pt,                  
    showspaces=false,                
    showstringspaces=false,
    showtabs=false,                  
}

\title{Pattern Recognition or Medical Knowledge? The Problem with Multiple-Choice Questions in Medicine}

\author{
  \textbf{Maxime Griot}\textsuperscript{1,2,3},
  \textbf{Jean Vanderdonckt}\textsuperscript{2},
  \textbf{Demet Yuksel}\textsuperscript{1,3},
  \textbf{Coralie Hemptinne}\textsuperscript{1,3}
\\
\\
  \textsuperscript{1}Institute of NeuroScience, Université catholique de Louvain, Brussels, Belgium\\
  \textsuperscript{2}Louvain Research Institute in Management and Organizations, Louvain-la-Neuve, Belgium\\
  \textsuperscript{3}Cliniques universitaires Saint-Luc, Brussels, Belgium
\\
  \small{
    \textbf{Correspondence:} \href{mailto:maxime.griot@uclouvain.be}{maxime.griot@uclouvain.be}
  }
}

\begin{document}
\maketitle
\begin{abstract}
Large Language Models (LLMs) such as ChatGPT demonstrate significant potential in the medical domain and are often evaluated using multiple-choice questions (MCQs) modeled on exams like the USMLE. However, such benchmarks may overestimate true clinical understanding by rewarding pattern recognition and test-taking heuristics. To investigate this, we created a fictional medical benchmark centered on an imaginary organ, the \textbf{Glianorex}, allowing us to separate memorized knowledge from reasoning ability. We generated textbooks and MCQs in English and French using leading LLMs, then evaluated proprietary, open-source, and domain-specific models in a zero-shot setting. Despite the fictional content, models achieved an average score of 64\%, while physicians scored only 27\%. Fine-tuned medical models outperformed base models in English but not in French. Ablation and interpretability analyses revealed that models frequently relied on shallow cues, test-taking strategies, and hallucinated reasoning to identify the correct choice. These results suggest that standard MCQ-based evaluations may not effectively measure clinical reasoning and highlight the need for more robust, clinically meaningful assessment methods for LLMs.
\end{abstract}
\section{Introduction}
\label{introduction}

Large Language Models (LLMs), such as ChatGPT, have demonstrated significant potential in the medical field, with studies evaluating their performance on tests originally designed for humans, including the United States Medical Licensing Examination (USMLE) \cite{jin_what_2020, pal_medmcqa_2022, jin_pubmedqa_2019, nori_capabilities_2023}. Furthermore, the domain-specific research shows that these models perform well on specialized medical exams in areas such as pediatrics, radiology, ophthalmology, plastic surgery, and oncology \cite{rydzewski_comparative_2024, bhayana_gpt-4_2023, barile_diagnostic_2024, mihalache_performance_2023, humar_chatgpt_2023}. The common reliance on MCQs in these assessments reflects their widespread use as a testing method for medical students around the globe \cite{al-wardy_assessment_2010}. 

However, while MCQs are easy to administer and grade, they have notable limitations, often promoting surface learning and pattern recognition over deep understanding \cite{veloski_patients_1999}. Despite their widespread use, few studies have addressed the potential issues unique to LLMs, such as their reliance on statistical patterns rather than genuine understanding. Notably, when trained on synthetic questions, \texttt{Meerkat-7b} outperformed its base model, \texttt{Mistral-7b}, on medical benchmarks by 18.6\%. This performance surpassed \texttt{Meditron-7b}, which improved by only 4\% despite being trained on a considerably larger, higher-quality clinical corpus \cite{kim_small_2024, chen_meditron-70b_2023}. This discrepancy highlights that extensive MCQ-based training can be more effective for benchmark performance than training on comprehensive medical content, raising concerns about the true depth of understanding being evaluated. This is further supported by more complex and realistic evaluations, such as patient interactions \cite{johri_evaluation_2025} or free-text questions \cite{arvidsson_chatgpt_2024}, which reveal that LLMs perform poorly compared to medical experts. 

These concerns are particularly relevant for LLMs, which are trained on large datasets that likely contain statistical patterns. This reliance can lead models to produce correct answers for incorrect reasons \cite{jin_hidden_2024}, such as identifying melanoma based on the presence of a ruler in an image \cite{narla_automated_2018}. While the limitations of MCQ-based medical benchmarks have begun to surface, recent work further underscores their fragility. The MedFuzz experiment \cite{ness_medfuzz_2024}, for example, showed that LLMs could be induced to provide incorrect answers by violating assumptions in the formulation of the questions. Likewise, replacing drug names with generic or branded alternatives led to performance drops of up to 10\% \cite{gallifant_language_2024}. Moreover, models struggle to identify when none of the options is correct or when the question cannot be answered due to missing information \cite{griot_large_2025}. Collectively, these findings suggest that strong performance on MCQs often reflects superficial pattern-matching rather than genuine medical understanding, as evidenced by the models' sensitivity to minor perturbations in otherwise familiar inputs.

Based on these findings, our work evaluates the performance of LLMs on novel medical concepts absent from the training data, thus investigating their ability to address questions about unfamiliar medical content. This approach investigates whether MCQ-based evaluations are primarily vulnerable to pattern variations or whether models can leverage test-taking strategies when encountering unfamiliar content. To achieve this, we developed a benchmark centered on a fictional organ, the Glianorex, designed to more effectively separate test-taking abilities from training data dependencies than previous studies have achieved.

\begin{figure*}[ht!]
\begin{center}
    \centerline{\includegraphics[width=\textwidth]{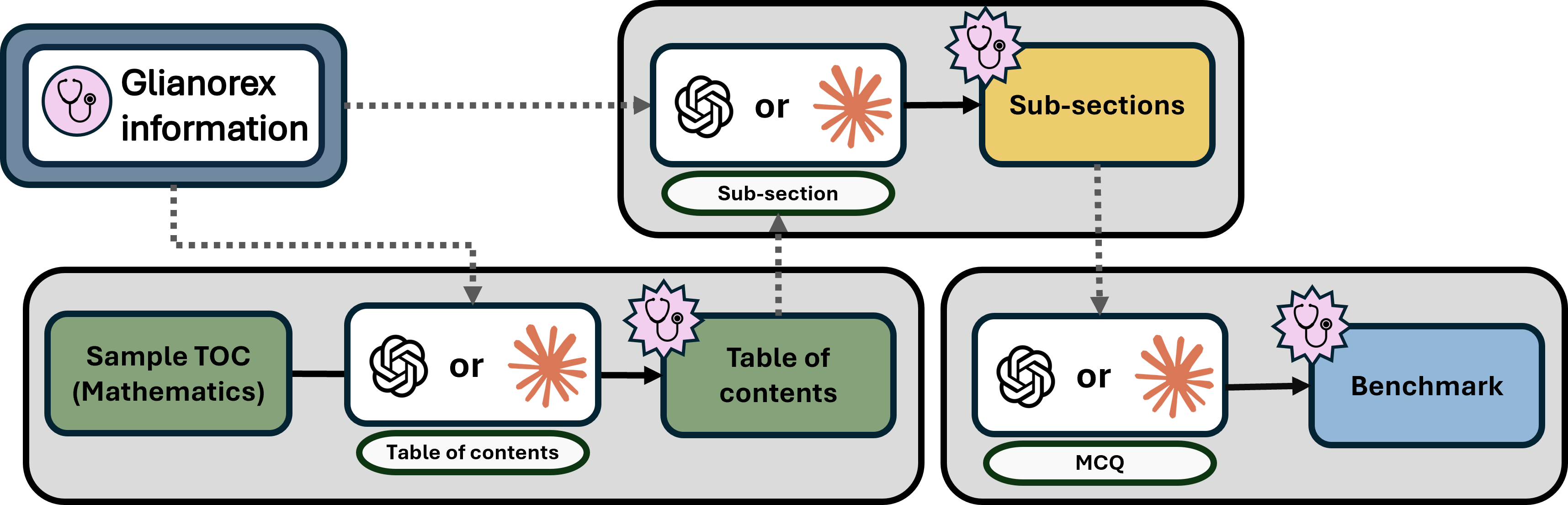}}
    \caption{Three-stage pipeline for benchmark generation: (1) Create a structured JSON table of contents using a math textbook template and Glianorex grounding data; (2) Generate detailed subsection content using hierarchical section titles and the same grounding data; (3) Iteratively generate questions for each subsection until at least 200 are created. Medical professionals perform quality assurance at every stage, marked by a stethoscope stamp.}
    \label{fig:methodology}
    \end{center}
\end{figure*}

\section{Related work}

The evaluation of medical knowledge and clinical skills remains an active research area, with new methods such as oral examinations and competency evaluations being proposed to better assess medical students and residents \cite{veloski_patients_1999, prediger_validation_2020, goins_use_2023}. Globally, medical evaluations are heavily based on MCQs, such as the USMLE in the United States, which significantly influences residency placements \cite{gauer_association_2017}. LLMs are similarly evaluated using MCQs to assess their medical knowledge. Google introduced MultiMedQA with its Med-PaLM model, which combines several existing medical benchmarks and has become a standard for evaluating medical proficiency in AI models \cite{singhal_large_2023, pal_openlifescienceaiopen_medical_llm_leaderboard_2024}. Recently, Google incorporated additional physician-led evaluations into its Med-Gemini model \cite{saab_capabilities_2024}. These multiple-choice items have raised concerns regarding their relevance for clinical use \cite{raji_its_2025}. MultiMedQA remains the most commonly reported benchmark to date and is composed of the following benchmarks:

\paragraph{MedQA-USMLE}
This subset of the MedQA dataset was sourced from the National Board of Medical Examiners (NBME), the organization responsible for the USMLE \cite{nbme_united_2024}. The dataset is composed of a total of 12723 questions, split into a training set of 10178 samples, a validation set of 1272 questions, and a test set of 1273 questions. The questions have 4 options with only one correct answer \cite{jin_what_2020}. Most questions present a clinical vignette and require the test-taker to apply clinical or foundational science knowledge to select the best answer.

\paragraph{MedMCQA}
The Multi-Subject Multi-Choice Dataset for Medical domain Question Answering is composed of 194k multiple choice questions obtained from the All India Institute of Medical Science (AIIMS) and National Eligibility cum Entrance Test Postgraduate (NEET PG) entrance examinations \cite{aiims_aiims_2024, nbems_neet_2024}. These questions are split into 3 subsets, one training subset composed of 183k samples, a validation subset of 4.18k samples, and a test subset comprising 6.15k samples, with the distinctive feature that the test subset omits the correct answers to prevent data leakage. The questions have 4 options each and can be either single or multiple choice. Most questions are straightforward knowledge-recall and do not use clinical vignettes.

\paragraph{PubMedQA}
This biomedical question answering dataset was created using PubMed \cite{nlm_pubmed_2024} article abstracts from which the authors derived a context with a question and a yes/maybe/no label. It comprises three subsets: an expert-annotated set of 1000 samples, an unlabeled set of 61.2k, and an automatically generated set of 211.3k. The generated samples are used to train models, while 500 samples of the expert-annotated subset are used to test the models. This benchmark was designed to evaluate the reasoning ability of models when presented with the abstract and a question related to this abstract \cite{jin_pubmedqa_2019}. 

\paragraph{MMLU-Medical}
The Massive Multitask Language Understanding dataset contains 57 tasks, of which six are used to assess medical knowledge (clinical knowledge, medical genetics, anatomy, professional medicine, college biology, and college medicine) \cite{hendrycks_measuring_2021}. These tasks were collected by students from publicly available online resources, including USMLE questions and undergraduate-level questions. The dataset contains 1,242 questions and is split into 30 for training, 123 for validation, and 1,089 for testing. The questions each have four options with only one correct answer and are a mix of clinical vignettes and recall questions.

\section{Methods}

To address these fundamental limitations in MCQ-based evaluation, we designed a novel benchmark to assess the relevance of MCQs for LLM evaluation using the process detailed in Figure~\ref{fig:methodology}. Our approach involved creating MCQs similar to those of the USMLE, but based on a fictional organ called the Glianorex. This process involved manual creation of grounding data by a volunteer physician who then prompted language models. The physician was instructed to write a brief overview of a fictional organ, including its name, the history of its discovery, anatomy, physiology (hormones and their role), histology (specific receptors), pathology (diseases associated with the Glianorex), and specific diagnostic techniques. 

\subsection{Dataset Construction}

\subsubsection{Knowledge}
To create diverse questions, the first step of the process was to augment the seed data using \texttt{GPT-4~Turbo} and \texttt{Claude 3.5 Sonnet} to generate additional content on the Glianorex. To generate a textbook, we first used the LLM to generate a table of contents in a standard JSON format with three levels of granularity: chapter, section, and subsection. After generation, manual verification was performed to validate the coherence and quality of the table of contents before proceeding with the textbook generation. The LLM was then used to generate each subsection independently. To improve coherence between different subsections, we provided the model with the grounding data and the complete table of contents. This process resulted in one textbook per model in English on the Glianorex detailing its history, physiology, anatomy, and pathology.

\subsubsection{Questions}
Based on these fictional textbooks, we used the same models separately to generate MCQs. The use of LLMs to generate questions based on textbooks was previously demonstrated to be an efficient and validated methodology, with \citet{kim_small_2024} showing significant improvements on various downstream benchmark tasks, including MedQA, MedMCQA, the USMLE sample test, and MMLU-Medical using this approach. This established precedent provides strong evidence that the quality of questions generated by state-of-the-art models would be sufficient for this study, even when applied to fictional medical content. For each model, these questions contained four choices with only one correct answer, adhering to a format similar to that of the USMLE to ensure uniformity. To facilitate the creation of these questions, we prompted models with the table of contents and a subsection from the textbook (see Table~\ref{prompt-table}). This approach guided both models to generate questions in a JSON format consistent with existing medical benchmarks.

\subsubsection{Multilingual}
To study the influence of language on test-taking abilities, we used the same models to translate the generated textbooks and questions using a simple one-shot prompt per subsection and question, asking the model to translate into French.

\subsubsection{Validation}
We recruited two physicians from our institution who completed at least one step of the USMLE in the past five years to assess the quality of the questions. They evaluated a random sample of 100 English questions on a 7-point Likert scale and answered them---without prior exposure to textbooks on the Glianorex---to establish an expert baseline. We conducted a keyword search for ``context'' across all questions to identify potential incompleteness. Finally, a physician manually verified the consistency of the Introduction, Anatomy, and Biochemistry chapters in both English and French \texttt{GPT-4 Turbo} generated textbooks to assess language quality, internal coherence, and translation quality.

\subsubsection{Synthetic Bias Mitigation}
Because our data-generation pipeline relies on LLMs, it is susceptible to synthetic biases. We therefore introduced several safeguards and human checkpoints:

\begin{enumerate}
    \item A physician authored the medical grounding information that was provided to the LLMs. This expert-verified context  aligned all subsections to a single source of truth.
    \item The entire experiment was replicated with two models of comparable capability (\texttt{GPT-4-Turbo} and \texttt{Claude 3.5 Sonnet}) using identical prompts and seed material to ensure that results were not model-specific.
    \item Before generation, the physician reviewed the table of contents to confirm its coherence and relevance. In addition, a sample of subsections was manually verified before proceeding with MCQ generation.
    \item To counter demographic bias and increase variability, we randomly specified gender and age parameters (ranging from 12 to 90 years) in 50\% of the prompts \cite{zack_assessing_2024}.
    \item For each subsection, we generated four questions with a temperature of~1 to produce diverse question variants.
    \item Answer options were shuffled so that the position of the correct choice was balanced across items.
    \item A sample of questions, textbook excerpts, and translations was audited by humans to verify quality and coherence.
\end{enumerate}

\subsection{Quantitative Analysis}

\subsubsection{Models}
To evaluate the performance of LLMs, we selected a diverse set of models, including both proprietary and open-weight options. We included commonly used foundational models, as reported in Table~\ref{models-table}. Additionally, we included two fine-tuned medical domain models based on \texttt{Mistral-7B-v0.1} to assess the influence of domain-specific training on this fictional benchmark. First, \texttt{internistai-7b-v0.2} (Apache 2.0), which was trained on a mixture of general data, medical textbooks, and MCQs, demonstrating improved performance on medical evaluations compared to its base model \cite{griot_impact_2024}. Second, \texttt{meerkat-7b-v1.0} (Creative Commons Attribution-NonCommercial 4.0), which was trained exclusively on MCQs, some of which were generated from medical textbooks \cite{kim_small_2024}. The latter training approach showed a significant performance increase in benchmarks using a relatively small amount of training data compared to continued pretraining on large medical datasets, as shown by \texttt{Meditron} and \texttt{PMC-LLaMA} \cite{chen_meditron-70b_2023, wu_pmc-llama_2024}.

\begin{table}[h!]
\begin{center}
\begin{small}
  \begin{tabular}{ll}
    \toprule
    \textbf{Model} & \textbf{License} \\
    \midrule
    {gpt-3.5-turbo-0125} & Proprietary \\
    {gpt-4-turbo-2024-04-09} & Proprietary  \\
    {gpt-4o-2024-05-13} & Proprietary  \\
    {01-ai/Yi-1.5-9B} & Apache 2.0  \\
    {01-ai/Yi-1.5-34B} & Apache 2.0  \\
    {mistralai/Mistral-7B-v0.1} & Apache 2.0 \\
    {mistralai/Mixtral-8x7B-v0.1} & Apache 2.0 \\
    {meta-llama/Meta-Llama-3-8B} & Llama 3 license  \\
    {meta-llama/Meta-Llama-3-70B} & Llama 3 license \\
    {Qwen/Qwen1.5-7B} & Tongyi Qianwen license \\
    {Qwen/Qwen1.5-32B} & Tongyi Qianwen license \\
    {Qwen/Qwen1.5-110B} & Tongyi Qianwen license \\
    \bottomrule
  \end{tabular}
  \end{small}
\end{center}
  \caption{\label{models-table}
  Foundational models \cite{openai_introducing_2022, openai_gpt-4_2023, openai_hello_2024, ai_yi_2024, mistral_models_2024, aimeta_llama_2024, bai_qwen_2023} included in the quantitative analysis.  }
  
\end{table}

\newpage

\subsubsection{Evaluation}

We evaluated all models using lm-evaluation-harness \cite{gao_framework_2023} in a zero-shot setting without additional training. The task followed the MedQA 4-option format, using a log-likelihood approach to measure accuracy. We calculated 95\% confidence intervals by multiplying the standard error of the mean by 1.96, assuming normal error distribution. We assessed statistical significance of accuracy against a random model using the cumulative distribution function of a binomial distribution.

A two-way analysis of variance (ANOVA) was conducted with model and benchmark subsets as independent variables, followed by Tukey's honestly significant difference (HSD) test for post-hoc pairwise comparisons of accuracy. We performed linear regression analysis to compare model accuracy on the Glianorex benchmark and MedQA-USMLE. Evaluations ran on a virtual machine with four NVIDIA A100 (80GB) GPUs on Microsoft Azure, with a total runtime of 10 hours including model download time.

\subsection{Interpretability and Ablation Analyses}
\label{sec:interpretability}

To examine how prompt structure affects performance and to understand the model's reasoning, we conducted ablation and qualitative studies with \texttt{DeepSeek-V3-0324} \cite{deepseek-ai_deepseek-aideepseek-v3-0324_2025} on the Glianorex benchmark in English and French. All generations were produced on a server with eight NVIDIA H200 (141GB) GPUs running \texttt{vLLM} \cite{kwon_efficient_2023} with greedy decoding (temperature = 0) to improve reproducibility.

\subsubsection{Prompting Parameters}
We evaluated four settings: \textbf{Zero-shot}, where the model sees the full question stem and answer choices and must select an answer directly; \textbf{Chain-of-Thought (CoT)}, which prompts the model to think step by step before the final answer; Zero-shot, \textbf{Answers-only (AO)}, where the question stem is removed and only answer options are provided; and \textbf{AO+CoT}, combining answers-only input with chain-of-thought reasoning.

\subsubsection{Analysis}
For each item and setting, we generated a single prediction and computed accuracy. Agreement between zero-shot and chain-of-thought predictions was assessed using Cohen’s $\kappa$. To test whether answer length influences selection, we compared the character length of the model's chosen option with the lengths of remaining alternatives.

Finally, we manually examined a subset of chain-of-thought traces, including both full questions and answers-only conditions, from correct and incorrect predictions in English.

\section{Results}

\subsection{Dataset}

The resulting fictional textbooks on the Glianorex were generated using the proposed structure with both \texttt{GPT-4 Turbo} and \texttt{Claude 3.5 Sonnet}. Each textbook contains detailed sections on the anatomy, physiology, biochemistry, pathology, and diagnostic tools related to the Glianorex. For both models, the textbooks were produced in English and French, each containing approximately 35,000 words. We then reused the subsections of the English textbooks to generate MCQs in English, followed by a translation step to obtain the same questions in French. The \texttt{GPT-4 Turbo} process resulted in 264 questions per language, while the \texttt{Claude 3.5 Sonnet} process produced 224 questions per language. For both models, examples of these questions (see Table~\ref{mcq-table} and Table~\ref{mcq-table-2}) included complex scenarios requiring multiple steps of reasoning. Each question adhered to a four-option format similar to MedQA-USMLE standards, with one correct answer.

\subsubsection{Internal consistency}
A partial data validation conducted by two physicians revealed no major flaws in the dataset. Their review of key textbook chapters identified minor inconsistencies that fell into two categories: contradictions and omissions. Contradictions involved discrepancies between subsections, such as slight variations in the described location of the Glianorex between the ``Proximity of the heart'' and ``Embryology and Development'' sections. Omissions occurred when relevant information appeared in only one subsection when it should have been present in others; for instance, the embryological origin of the Glianorex as \emph{splanchnopleure} was mentioned in the ``Vascular Supply'' section but absent from ``Embryology and Development.'' While these inconsistencies were subtle and required extensive cross-referencing to detect, they did not compromise the overall integrity of the content. 

\subsubsection{Language}
Cross-linguistic analysis demonstrated structural and content consistency across languages, with only negligible variations in French abbreviation conventions. Quality assessment of a 100-question English sample by two physicians yielded high scores (6.94 and 6.86 out of 7), indicating quality comparable to board-examination standards. Manual verification across both languages identified only eight incomplete questions (four per language, $<1\%$ of total) that required additional context to be answered.

\subsection{Evaluations}

\subsubsection{General Results}

\begin{figure}[t!]
\begin{center}
    \centerline{\includegraphics[width=0.5\textwidth]{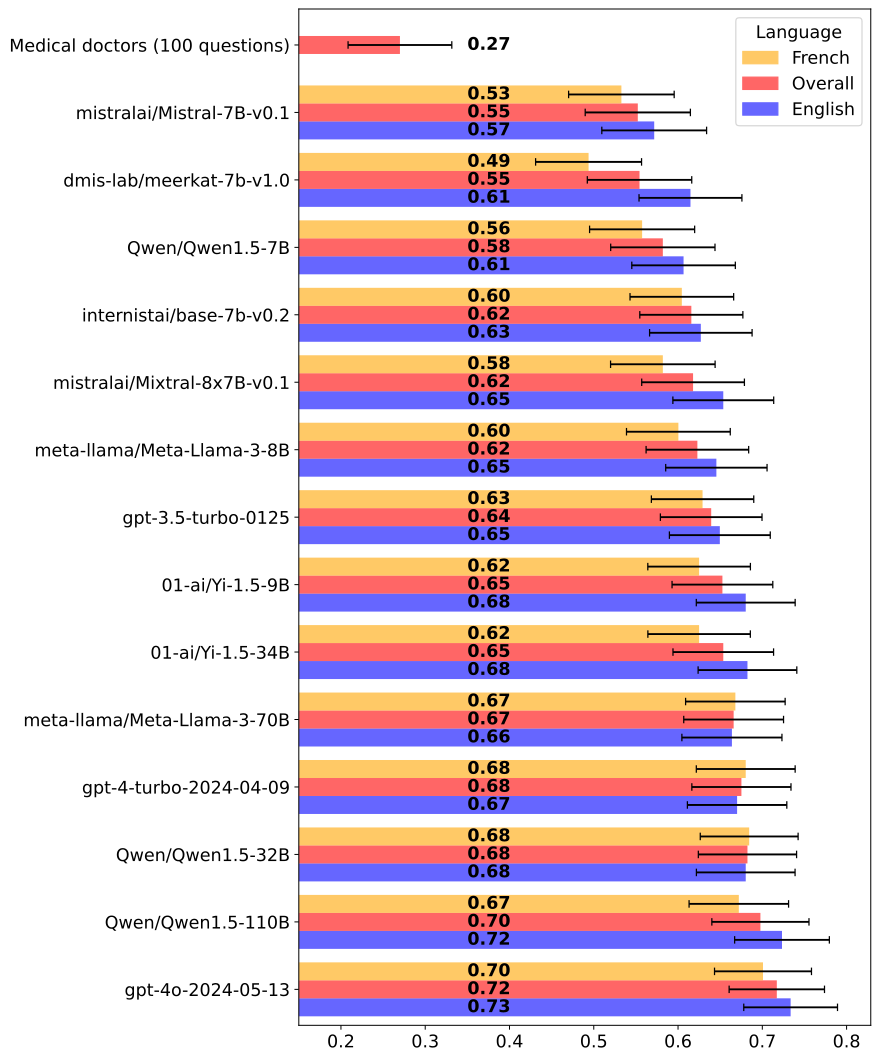}}
    \caption{Accuracy of the evaluated models on the synthetic benchmark with 95\% confidence intervals. Scores are presented separately for English and French, illustrating that most models achieve higher accuracy in English compared to French. Additionally, we include the performance of medical doctors evaluated on a subset of 100 English questions as a human reference.}
    \label{fig:accuracy}
    \end{center}
\end{figure}

All models achieved relatively high scores, averaging 63.8\%, as illustrated in Figure~\ref{fig:accuracy}. To place this score in perspective, the physicians each obtained 27\%, which is within the expected range for random answering. The physicians noted that the questions relied heavily on fictional terminology and concepts, and a substantial portion focused on information recall, leading them to resort to guesswork rather than applying their medical expertise to formulate responses. 

A statistically significant difference was observed between the top-performing models and the lowest-performing models, as shown in Table~\ref{merged-pvalues-table}. The performance differences when isolating languages were also significant and occurred more frequently in English, as shown in Table~\ref{tab:english_compare} and Table~\ref{tab:french_compare}. We also calculated Cohen's d between all model pairs, which revealed a range of effect sizes, indicating varying degrees of performance differences between the models \cite{cohen_statistical_2013}. Most of the comparisons show very small or negligible effect sizes, with many pairs having a Cohen's d close to 0, as shown in Table~\ref{tab:cohend}. For instance, pairs such as \texttt{Yi-1.5-34B} -- \texttt{Yi-1.5-9B} ($d = 0.002$) and \texttt{Yi-1.5-34B} -- \texttt{gpt-3.5-turbo-0125} ($d = 0.030$) suggest negligible differences. This pattern is consistent across most pairs, indicating that the models' performances are closely aligned. 

However, some pairs demonstrate more noticeable differences, such as \texttt{meerkat-7b-v1.0} -- \texttt{gpt-4o-2024-05-13} ($d = 0.343$) and \texttt{gpt-4-turbo-2024-04-09} -- \texttt{Mistral-7B-v0.1} ($d = 0.254$), suggesting a measurable effect. Overall, the analysis reveals that while some variations exist, the effect sizes for most model comparisons are small. Additionally, the average score for English questions was 65.7\%, whereas French averaged 61.8\%.

\begin{table}[!t]
\begin{center}
\begin{footnotesize}
\resizebox{\columnwidth}{!}{
\begin{tabular}{lllll}
\toprule
 & \rotatebox{90}{Qwen/Qwen1.5-110B} & \rotatebox{90}{dmis-lab/meerkat-7b-v1.0} & \rotatebox{90}{gpt-4o-2024-05-13} & \rotatebox{90}{mistralai/Mistral-7B-v0.1} \\
\midrule
Qwen/Qwen1.5-7B & * &   & ** &   \\
meta-llama/Meta-Llama-3-70B &   & * &   & * \\
Qwen/Qwen1.5-32B &   & ** &   & ** \\
Qwen/Qwen1.5-110B &   & ** &   & *** \\
gpt-4-turbo-2024-04-09 &  &  *&   & * \\
gpt-4o-2024-05-13 &  &  ****&   & **** \\
\bottomrule
\end{tabular}
}
\end{footnotesize}
\end{center}
\caption{\label{merged-pvalues-table}Statistical significance of the performance differences between models (* p $<$ 0.05, ** p $<$ 0.01, *** p $<$ 0.001, and **** p $<$ 0.0001). }
\end{table}

\paragraph{Finetuned Models}
The \texttt{internistai-7b-v0.2} and \texttt{meerkat-7b-v1.0} models demonstrated enhanced English performance relative to their base model, \texttt{Mistral-7B-v0.1}. However, this improvement was not replicated in French, suggesting that domain-specific training enhances performance in the target language, but the absence of multilingual data during continued training may diminish performance in other languages.

\subsubsection{Cross-Benchmark Analysis}

\begin{figure}[!t]
\begin{center}
\centerline{\includegraphics[width=0.5\textwidth]{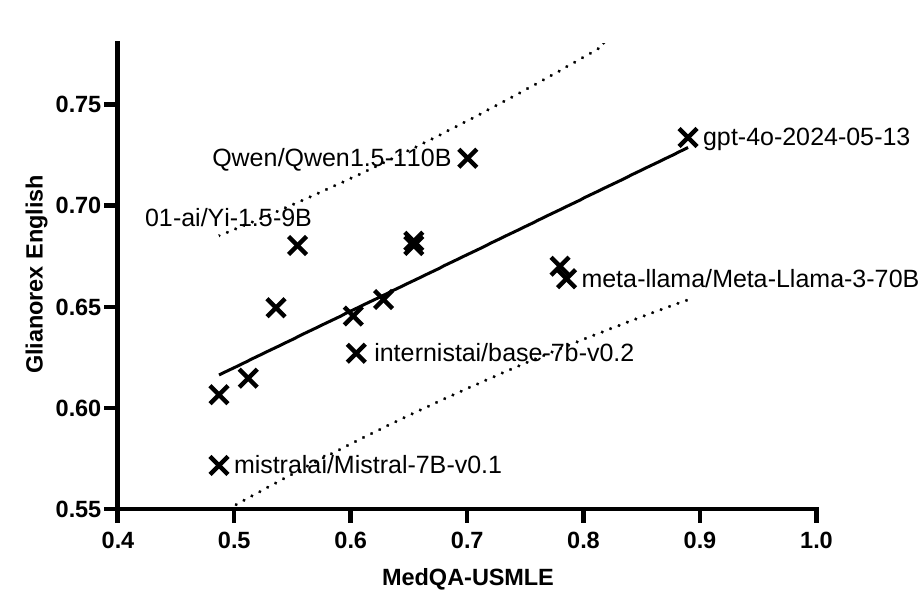}}
    \caption{Linear regression analysis comparing MedQA-USMLE four-option scores and Glianorex English scores, shown with 95\% prediction bands.}
    \label{fig:regression}
    \end{center}
\vskip -0.2in
\end{figure}

We conducted a linear regression analysis to examine the relationship between performance on the MedQA-USMLE four-option benchmark and accuracy on the Glianorex English subset, as illustrated in Figure~\ref{fig:regression}. The results revealed a statistically significant correlation ($p < 0.01$) between these two metrics ($R^{2} = 0.5952$). This correlation, combined with the observed relationship between model size and performance, suggests that improvements in medical benchmarks may be partially attributable to enhanced pattern recognition capabilities.

\FloatBarrier

\subsection{Interpretability}

\subsubsection{Input and Reasoning Ablations}

In the zero-shot configuration, \texttt{DeepSeek-V3-0324} achieved an average accuracy of 70.1\%, consistent with the top-performing models previously evaluated. Chain-of-thought prompting yielded only marginal improvement (70.3\%), indicating limited effectiveness of explicit reasoning prompts for this task (Table~\ref{tab:ablation}). Cohen's $\kappa$ analysis confirmed substantial agreement between zero-shot and chain-of-thought prompting approaches ($\kappa = 0.681$ for English, $\kappa = 0.653$ for French).

When prompted exclusively with answer choices (Answers-only setting), the model achieved an average accuracy of 46.5\%. While this performance significantly exceeds random chance (25\%), it remains substantially below full-prompt performance. This intermediate accuracy level supports the hypothesis proposed by \citet{balepur_artifacts_2024}, suggesting that large language models employ meta-strategies such as question inference and surface-level shortcuts. Furthermore, we analyzed the relative length of answers selected by the model compared to other available choices and found no statistically significant differences ($p > 0.05$).

\begin{table}[!t]
\begin{center}
\begin{footnotesize}
\resizebox{\columnwidth}{!}{
\begin{tabular}{llllll}
\toprule
 & zero-shot & CoT & AO + zero-shot & AO + CoT \\
\midrule
English & 0.705 & \textbf{0.717} & 0.457 & 0.473 \\
French &  \textbf{0.697} & 0.689 & 0.473 & 0.438 \\
Average & 0.701 &  \textbf{0.703} & 0.465 & 0.455 \\
\bottomrule
\end{tabular}
}
\end{footnotesize}
\end{center}
\caption{\label{tab:ablation}Accuracy of \texttt{DeepSeek-V3-0324} on the benchmark using four evaluation configurations. AO (Answers Only) prompts only with the choices, the question stem is removed entirely.}
\end{table}

\subsubsection{Qualitative Analysis}

Manual qualitative examination of chain-of-thought explanations identified three recurring patterns employed by the model: hallucinations, generalized medical assumptions, and explicit test-taking strategies. For each pattern, we describe common characteristics and present examples as generated by \texttt{DeepSeek-V3-0324}.

\subsubsection{Hallucinations}
The model frequently generated fabricated knowledge based on question and answer content. This pattern predominantly resulted in incorrect responses but occasionally led to correct answers. For example, the model hallucinated information regarding the optimal imaging technique for a fictional disease, which guided the model toward an incorrect answer.

\begin{promptbox}[Hallucination]
\small
Glianorex Imagery Sonography (GIS) is the most specific and helpful diagnostic tool for confirming autoimmune Glianorexiditis. This imaging modality allows for direct visualization of glianorex tissue inflammation and damage, which is critical for diagnosis.
\end{promptbox}

\subsubsection{Medical Assumptions}
The model explicitly applied characteristics of real autoimmune diseases to the fictional conditions presented in the benchmark, occasionally yielding correct inferences but frequently causing inaccuracies. For instance, the model erroneously assumed that genetic risk factors associated with known autoimmune disorders would similarly apply to the fictional condition Glianorexiditis.

\begin{promptbox}[General medical principles]
\small
The question involves a fictional condition ("Glianorex degeneration") and diagnostic test ("Glianorex Imagery Sonography (GIS)"), so the answer must be inferred from the context of the question and general medical principles
\end{promptbox}

\subsubsection{Test-Taking Strategies}
We identified explicit heuristic approaches, including the selection of highly specific answers and the preference for answers that structurally resemble typical examination formulations. This strategy is also observed, albeit to a lesser extent, in human test-takers, such as avoiding answers containing absolute terms like "always" or "never," which tend to be incorrect in medical contexts.

\begin{promptbox}[Specificity]
\small
In exams, highly specific answers ("biotransplant," "modulators," "re-equilibration") are often correct when other choices are generic.
\end{promptbox}

The model also explicitly recognized and exploited answer constructions that it perceived as characteristic of examination environments.

\begin{promptbox}[Test construction]
\small
D stands out as the most "constructed" correct answer in a medical context, resembling how hypothetical disorders are framed in exams. (While all options contain questionable terms, D is the most logically structured and aligns best with how exam questions are typically designed - tying a novel mechanism to a targeted treatment.)
\end{promptbox}

\section{Discussion}
\label{discussion}

\subsection{Evaluation Implications}

The results of this study highlight several insights into the capabilities and limitations of LLMs in handling medical MCQs. Despite the novelty and complexity of the fictional organ, all evaluated models achieved high scores on the MCQs generated for this material. However, physicians who attempted to answer a random subset of the benchmark were unable to perform better than chance. This finding suggests that LLMs are adept at recognizing patterns and applying test-taking strategies, even in unfamiliar contexts.

\subsubsection{Benchmarking}
The consistently high performance across various foundational models in English, regardless of their architecture, size, or specialization, indicates that traditional MCQ-based benchmarks may inadequately assess LLMs' medical knowledge and clinical reasoning skills. These benchmarks appear to test pattern identification and association abilities rather than genuine material comprehension. Consequently, relying on MCQs to evaluate LLMs in medical and other specialized domains might overestimate their actual capabilities. This finding aligns with research demonstrating that models become less reliable as they scale up \cite{zhou_larger_2024}. Using adversarial benchmarks like the one introduced in this study could help identify reliability reductions during development.

\subsubsection{Training}
The superior performance of fine-tuned models \texttt{internistai/base-7b-v0.2} and \texttt{dmis-lab/meerkat-7b-v1.0} over the foundational model \texttt{mistralai/Mistral-7B-v0.1} underscores the impact of task-specific and domain-specific training on LLM capabilities. Both models were trained on medical MCQs---with Meerkat trained exclusively on MCQs---raising the question of whether the improvement stems from enhanced test-taking skills or greater medical-domain knowledge. Previous research shows that task improvements in models trained on additional medical data disappear after prompt optimization \cite{jeong_medical_2024}, suggesting that additional training may target evaluation methodology to improve accuracy rather than enhancing medical capabilities. This aligns with findings that models need specific training on extraction tasks to leverage their internal knowledge, which may explain the gains observed after additional MCQ training \cite{allen-zhu_physics_2024}.

\subsection{Medical Implications}

Current medical evaluation standards may not accurately reflect LLMs' capabilities in the medical domain, raising significant concerns about their safety and clinical implications in real-world settings. Performance claims based on MCQs could misrepresent these models' actual capabilities, creating false trust that might endanger patients who rely on these systems instead of consulting physicians, and physicians who implement them for clinical decision support.

Such claims could also undermine trust within the medical community, which has already expressed skepticism regarding LLM applications in medicine \cite{marks_ai_2023, flanagin_nonhuman_2023}. Misrepresenting the medical capabilities and usefulness of these models may lead physicians to view AI as a commercial selling point rather than a tool for real progress, potentially hindering AI adoption and limiting opportunities for multidisciplinary teams to develop clinically relevant models.

The integration of LLMs in clinical settings poses significant patient safety risks, especially given the time constraints faced by clinicians. Previous work by \citet{liu_artificial_2024} demonstrates a significant dose-response association between AI usage and burnout for radiologists, as well as increased post-processing, which can be attributed partly to added validation time. Given these findings and the high risk posed by these models, we believe it is unrealistic to expect clinicians to read chain-of-thought reasoning sections to ensure response validity, and therefore believe that models should provide trustworthy responses in zero-shot settings.

\subsection{Recommendations}
We recommend including medical professionals in model evaluation and urge developers to exercise greater caution when making claims based on MCQ-based benchmarks. Similar to medical devices and drugs, models should undergo clinical trials to ensure safety and demonstrate patient benefits over current practices \cite{widner_lessons_2023}. This requires a paradigm shift toward answering concrete questions such as "Does the use of model X to recommend parenteral nutrition reduce mortality in hospitalized patients with neck cancer?" instead of the current approach of assessing medical capabilities, a task that lacks both proper definition and clinical practice relevance.

\subsection{Alternatives}
Nevertheless, there is a need for automated and standardized evaluations to guide development. More advanced methodologies that do not rely solely on MCQs have been proposed, including case-based reasoning scenarios requiring intermediate physiological explanations, similar to those in the French ECN exams \cite{cng_sante_epreuve_2021}. Additionally, key-feature problems, where clinicians must identify critical decision points within complex clinical scenarios and prioritize among multiple correct options, can offer deeper insights into model capabilities \cite{bordage_key-features_2018}. Open-ended questions evaluated through rubric-based approaches, combining LLM-as-judge assessments with expert verification, can further enhance evaluation validity \cite{zheng_judging_2023}. Finally, simulated clinical environments, such as the adaptive questioning and diagnostic refinement demonstrated in AI Hospital by \citet{fan_ai_2025}, present complex, dynamic settings to assess and refine LLM performance more accurately for safe and effective clinical application.

\section{Conclusion}

This study demonstrates that LLMs can achieve high scores on MCQs built around fictional medical knowledge without prior exposure to the content. By creating a fictional gland, the Glianorex, and generating comprehensive textbooks with related MCQs, we partially isolated the models' reasoning capabilities from memorized real-world data. Results show that models of different architectures, sizes, and specializations outperform physicians on this benchmark, suggesting that pattern recognition and test-taking strategies may play a larger role for LLMs than for humans.

Our findings call into question the effectiveness of current MCQ-based benchmarks for evaluating LLMs' clinical knowledge and reasoning abilities. This study highlights the need for more robust evaluation methods that better assess the true understanding and reasoning capabilities of LLMs in the medical domain. Future research should explore alternative evaluation methods beyond current MCQs to provide more accurate assessments of LLMs' capabilities in medicine and other specialized fields.

\section{Limitations}

\paragraph{Knowledge coherence}
Independent generation of subsections could result in inconsistencies or contradictions within the text, potentially creating questions with multiple plausible correct answers depending on the chapter context provided during question generation. We performed a partial coherence check on the generated textbook to ensure content plausibility and identified few inconsistencies and contradictions. This partial check does not guarantee the absence of major errors; however, since LLMs had no prior exposure to this fictional knowledge, inconsistencies between independent subsections should not affect their ability to answer appropriately.

\paragraph{Synthetic biases}
Although we took steps to reduce synthetic biases intrinsic to our methodology, some may persist in the dataset. For example, grounding information authored by a single clinician inevitably reflects that clinician's biases, which could partly account for the models' tendency to leverage common medical patterns when answering questions on fictional content. Nevertheless, these residual biases are unlikely to fully explain the observed performance gap, especially given the low scores achieved by two physicians. To corroborate our findings, future work should evaluate models on multiple-choice questions covering clinical knowledge published after the training cut-off date.

\section{Acknowledgements}

This work was supported by the Fondation Saint-Luc grant number 467E and the Fédération Wallonie-Bruxelles through the Fond Spécial de Recherche of Université catholique de Louvain.

\bibliography{acl_latex}

\newpage
\clearpage

\appendix

\section{Reproducibility}

\subsection{Code}
\label{sec:code}

\paragraph{lm-evaluation-harness}
The main branch of lm-eval-harness contains the $glianorex$, $glianorex\_en$, and $glianorex\_fr$ tasks under the MIT license \url{https://github.com/EleutherAI/lm-evaluation-harness/pull/1867}.

\paragraph{Synthetic generation}
The code used to generate the synthetic dataset and multiple choice questions is available under the MIT license on GitHub and contains the textbooks generated for this study. DOI: \href{https://doi.org/10.5281/zenodo.15496631}{10.5281/zenodo.15496631}. 

\paragraph{GPT evaluation}
Due to the limitations of lm-evaluation-harness with OpenAI models, we had to write OpenAI-specific code to evaluate the models available under the MIT license on GitHub. DOI: \href{https://doi.org/10.5281/zenodo.15496636}{10.5281/zenodo.15496636}.

\subsection{Parameters}

The API parameters used to generate the book, translate, and generate multiple-choice questions are the default parameters as shown in Table \ref{tab:parameters}.

\begin{table}[!ht]

\centering
\begin{tabular}{ll}
\toprule
Parameter & Value \\
\midrule
frequency\_penalty & 0 \\
n & 1 \\
presence\_penalty & 0 \\
temperature & 1.0 \\
top\_p & 1.0 \\
\bottomrule
\end{tabular}
\caption{API parameters}
\label{tab:parameters}
\end{table}

\newpage
\pagebreak
\newpage

\begin{table*}[!ht]
\begin{center}
\begin{small}
  \begin{tabular}{p{0.8cm}p{\linewidth - 0.8cm - 4\tabcolsep}}
    \toprule
    \textbf{Role} & \textbf{Content} \\
    \midrule
    System & You are a helpful assistant helping generate knowledge on a fictional gland and its associated diseases. You are tasked with transforming the existing text to generate variations to help learn the content. \\
    \midrule
    User & You are given some context and a table of contents to help:
    
    \textbf{TABLE OF CONTENTS}
    
    Query: Generate a very complicated multiple-choice question requiring multiple steps of reasoning with 4 options, these are not reading questions but a test to ensure the student understands and knows the content. Here is an example json output, match this format:

    \begin{verbatim}
```json
{
  "question": "The question",
  "choices": ["(A) Choice A", 
    "(B) Choice B", 
    "(C) Choice C", 
    "(D) Choice D"],
  "solution": "(D) Choice D"
}
```
\end{verbatim}
    
    Text:
    \textbf{TEXTBOOK PARAGRAPH}\\
    \bottomrule
  \end{tabular}
    \end{small}
\end{center}
  \caption{\label{prompt-table}
  The prompt used to generate multiple-choice questions is based on a subset of the textbook. The prompt template contains two variables \textbf{TABLE OF CONTENTS} and \textbf{TEXTBOOK PARAGRAPH}, which are respectively replaced with the table of contents of the textbook and a random paragraph from the textbook to provide context to the model.}
 \end{table*}

\subsection{Evaluation}

To evaluate the open-weight models, we used lm-evaluation-harness, which includes the Glianorex tasks. For any pre-trained model hosted on HuggingFace, replace \texttt{MODEL} with the path of the model and run the following command:
\lstset{style=mystyle}
\begin{lstlisting}[language=Bash]
lm_eval --model hf 
  --model_args pretrained=MODEL,dtype="bfloat16",parallelize=True 
  --tasks glianorex_en,glianorex_fr 
  --batch_size 32 
  --log_samples 
  --output_path /tmp/results
\end{lstlisting}

The hardware needed depends on the size of the model; we recommend at least 4 NVIDIA A100 80GB to evaluate models of 70 billion parameters. Reducing $batch\_size$ can help reduce memory requirements. The standalone questions dataset can be found under the MIT license on HuggingFace. DOI: \href{https://doi.org/10.57967/hf/2344}{10.57967/hf/2344}.

\subsection{Human Annotation}

\begin{figure*}[!ht]
\begin{center}
\centerline{\includegraphics[width=0.7\textwidth]{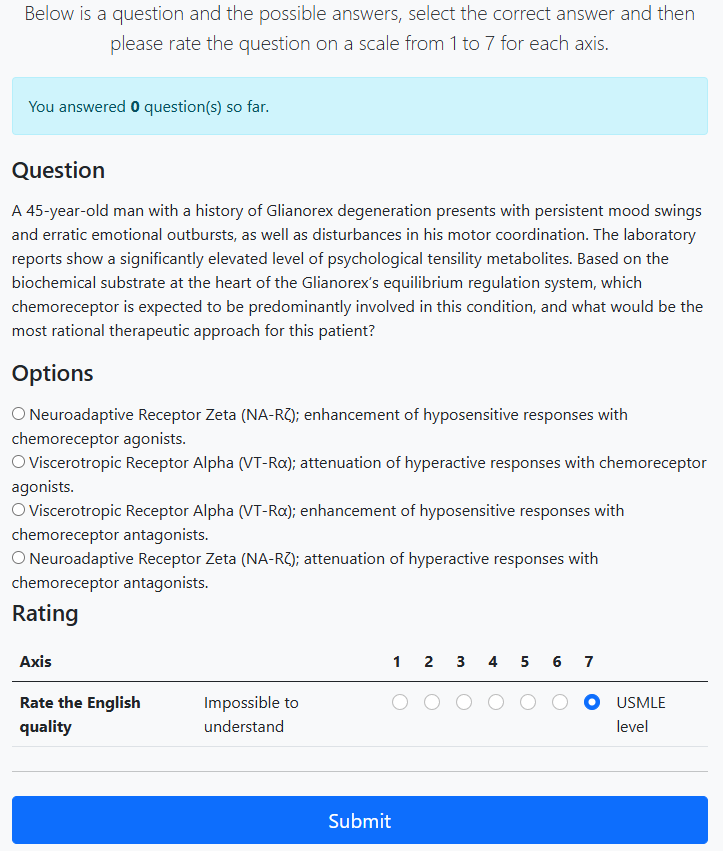}}
    \caption{User interface presented to human test takers.}
    \label{fig:survey}
    \end{center}
\end{figure*}

Human annotators used a website designed for this experiment that presented 100 questions in a randomized order. Each question had 4 options, only one of which was correct. The annotator had to select one of the options and then rate on a 7-point Likert scale the English quality, with 1 being ``Impossible to understand'' and 7 being ``USMLE level'' as shown in Figure~\ref{fig:survey}.

\newpage

\onecolumn

\section{Additional Results}

\begin{table}[!ht]
\centering
\begin{tabular}{lllll}
\toprule
 \multicolumn{1}{c}{Model} & \multicolumn{2}{c}{English} & \multicolumn{2}{c}{French} \\
\cmidrule(lr){2-3} \cmidrule(lr){4-5}
 & GPT & Claude & GPT & Claude \\
\midrule
01-ai/Yi-1.5-34B & \textbf{0.70} & 0.66 &0.61 & \textbf{0.64} \\
01-ai/Yi-1.5-9B & \textbf{0.69} & 0.67 &\textbf{0.62} & 0.62 \\
dmis-lab/meerkat-7b-v1.0 & \textbf{0.66} & 0.57 &0.49 & \textbf{0.50} \\
gpt-3.5-turbo-0125 & \textbf{0.69} & 0.60 &\textbf{0.64} & 0.61 \\
gpt-4-turbo-2024-04-09 & 0.65 & \textbf{0.69} &\textbf{0.68} & 0.68 \\
gpt-4o-2024-05-13 & \textbf{0.74} & 0.72 &0.69 & \textbf{0.71} \\
internistai/base-7b-v0.2 & \textbf{0.64} & 0.61 &\textbf{0.61} & 0.59 \\
meta-llama/Meta-Llama-3-70B & 0.66 & \textbf{0.67} &0.65 & \textbf{0.69} \\
meta-llama/Meta-Llama-3-8B & 0.64 & \textbf{0.65} &0.59 & \textbf{0.61} \\
mistralai/Mistral-7B-v0.1 & \textbf{0.59} & 0.54 &\textbf{0.56} & 0.50 \\
mistralai/Mixtral-8x7B-v0.1 & \textbf{0.68} & 0.62 &\textbf{0.59} & 0.58 \\
Qwen/Qwen1.5-110B & 0.72 & \textbf{0.73} &\textbf{0.67} & 0.67 \\
Qwen/Qwen1.5-32B & 0.67 & \textbf{0.70} &0.65 & \textbf{0.73} \\
Qwen/Qwen1.5-7B & \textbf{0.62} & 0.59 &0.53 & \textbf{0.58} \\
\bottomrule
\end{tabular}
\caption{Comparison of model performances depending on language and the model used to generate questions. Bolded values indicate the highest accuracy for the current language. The models compared are gpt-4-turbo-2024-04-09 (GPT) and Claude Sonnet 3.5 (Claude).}
\label{tab:model_comparison}
\end{table}

\begin{table}[!ht]
\centering
\begin{tabular}{llll}
\toprule
 & \rotatebox{45}{Qwen1.5-110B}& \rotatebox{45}{gpt-4o-2024-05-13}& \rotatebox{45}{Mistral-7B-v0.1}\\
\midrule
Qwen1.5-7B & * & ** &   \\
 meerkat-7b-v1.0& *& *&\\
Meta-Llama-3-70B &   &   & * \\
Yi-1.5-9B &   &   & * \\
Yi-1.5-34B &   &   & * \\
Qwen1.5-32B &   &   & * \\
Qwen1.5-110B &   &   & ** \\
internistai/base-7b-v0.2 & * & * &   \\
gpt-4-turbo-2024-04-09 &  &   & * \\
gpt-4o-2024-05-13 &  &   & *** \\
\bottomrule
\end{tabular}
\caption{Statistical significance of the performance differences in English between models (* p $<$ 0.05, ** p $<$ 0.01, *** p $<$ 0.001, and **** p $<$ 0.0001).}
\label{tab:english_compare}
\end{table}

\begin{table}[!ht]
\centering
\begin{tabular}{lll}
\toprule
 & meerkat-7b-v1.0& Mistral-7B-v0.1\\
\midrule
Meta-Llama-3-70B & * &   \\
Qwen1.5-32B & ** &   \\
Qwen1.5-110B & * &   \\
gpt-4o-2024-05-13 &  ***& * \\
 gpt-4-turbo-2024-04-09& **&\\
 \bottomrule

\end{tabular}
\caption{Statistical significance of the performance differences in French between models (* p $<$ 0.05, ** p $<$ 0.01, *** p $<$ 0.001, and **** p $<$ 0.0001).}
\label{tab:french_compare}
\end{table}

\begin{table}[!htbp]
  \centering
  \begin{tabular}{p{0.95\linewidth}}
    \toprule
    \textbf{Content} \\
    \midrule
A 45 year-old male who works night shifts is hospitalized following an episode of severe mood swings and physical tremors. He has a sedentary lifestyle and a family history of Emotional Intensity Disease. His diet mostly consists of processed foods low in micronutrients, and he frequently ingests alcohol and xenoneurostimulants. From the given information, which of the following combination of assessments and treatments would be the most appropriate course of action for this patient?

\\

(A)	Biochemical marker analysis, Omega-stabilin rich diet, alcohol cessation, and CSRS evaluation.

(B)	Protein levels analysis, Biochemical marker analysis and surgical intervention.

\textbf{(C)	Biochemical marker analysis, Nutrilyte Complex supplementation, personalised exercise plan, alcohol cessation, circadian alignment strategy, and adoption of stress management techniques.}

(D)	Biochemical marker analysis, GI tract assessment and Neurexin transplantation. \\
\midrule
Un homme de 35 ans est diagnostiqué avec la Maladie d'Intensité Émotionnelle et se plaint de fatigue diurne sévère et de sautes d'humeur. Ses enregistrements polysomnographiques montrent des signes d'une architecture du sommeil perturbée, y compris une paralysie du sommeil. Il rapporte une émotivité au réveil et un sommeil non réparateur. Ses échantillons de sérum montrent un niveau élevé de Somnolabilin nocturne et un schéma de sécrétion de Nocturnin perturbé. Compte tenu de ces résultats, quelle méthodologie a probablement été utilisée pour diagnostiquer son état, quelle hormone est probablement associée à sa perturbation du sommeil et à son atonie physique, et quelle pourrait être une stratégie de traitement possible ?

\\
(A)	Diagnostic avec la Chrono-Enzyme-Linked Immunosorbent Spectroscopy (C-ELIS) d'Elara-Mendoza, l'hormone Nocturnin devrait être associée à ses symptômes et des interventions pharmaceutiques ciblant la synthèse de Nocturnin comme traitement.

(B)	Diagnostic avec des essais d'électrovalence synaptique, l'hormone Somnolabilin devrait être associée à ses symptômes et des modifications du mode de vie comme traitement.

\textbf{(C)	Diagnostic avec la Chrono-Enzyme-Linked Immunosorbent Spectroscopy (C-ELIS) d'Elara-Mendoza, l'hormone Somnolabilin devrait être associée à ses symptômes et des interventions pharmaceutiques ciblant la synthèse de Somnolabilin comme traitement.}

(D)	Diagnostic avec des enregistrements polysomnographiques, l'hormone Nocturnin devrait être associée à ses symptômes et la chronothérapie comme traitement. \\

    \bottomrule
  \end{tabular}
    \caption{Example of clinical vignette questions in English and French generated by GPT-4 Turbo on a random paragraph of the textbook. The correct answer is shown in bold.}
  \label{mcq-table}
\end{table}

\begin{table}[!ht]
  \centering
  \begin{tabular}{p{0.95\linewidth}}
    \toprule
    \textbf{Content} \\
    \midrule
Considering the detailed anatomy and vascular supply of the Glianorex, which of the following processes best describes how the Glianorex modulates its endocrine functions in response to emotional stimuli?

\\

(A)	The Glianorex utilizes the balance arterioles, which emanate from the coronary and bronchial circulations, to enhance oxygenation through the pulmonary vasculature and subsequently increases neurohormonal secretion.

(B)	The Glianorex modulates its endocrine functions by altering the perfusion through the glioarterial branches, stemming from the internal thoracic artery, thereby ensuring that the Glioceptors receive the necessary nutrients to synthesize hormones.

(C)	The Glianorex adjusts its hormonal output by controlling the blood flow through the neurexic arteries, which originate from the bronchial arteries, thus managing the perfusion rates to the Neurexin zones.

\textbf{(D)	The Glianorex relies on pre-capillary sphincters and post-capillary venules equipped with smooth muscle fibers to regulate oxygenation of its parenchyma, which reflexively adjusts the organ's hormone secretion in alignment with neurohormonal stimuli.} \\
\midrule
Quelle est la séquence correcte des voies nerveuses et leurs fonctions principales associées au sein du réseau du Glianorex, partant de la détection du stimulus émotionnel jusqu'à la sortie hormonale finale ?

\\
(A)	Détection via les Gliocepteurs -> Intégration par les Globuli Emotoafférents -> Traitement par les Ganglions Sentirex -> Sortie hormonale avec Equilibron et Neurostabilin

\textbf{(B)	Détection via les Gliocepteurs -> Traitement par les Ganglions Sentirex -> Sortie hormonale avec Equilibron et Neurostabilin médiée par les Psychoneurexines -> Modulation synaptique par le Synaptome Séraphique}

(C)	Détection via les Globuli Emotoafférents -> Traitement par les Ganglions Sentirex -> Sortie hormonale avec Equilibron et Neurostabilin médiée par la Voie Gliopathique Primordiale -> Modulation de la sensibilité des Gliocepteurs par le Synaptome Séraphique

(D)	Détection via les Gliocepteurs -> Intégration par les Psychoneurexines -> Traitement par les Ganglions Sentirex -> Sortie hormonale avec le Synaptome et l'Alectorol \\

    \bottomrule
  \end{tabular}
  \caption{Example of recall questions in English and French generated by GPT-4 Turbo on a random paragraph of the textbook. The correct answer is shown in bold.}
  \label{mcq-table-2}
\end{table}

\FloatBarrier

\begin{longtable}{llr}
\toprule
Model 1 & Model 2 & Cohen's d \\
\midrule
\endfirsthead
\toprule
Model 1 & Model 2 & Cohen's d \\
\midrule
\endhead
\midrule
\multicolumn{3}{r}{Continued on next page} \\
\midrule
\endfoot
\endlastfoot
01-ai/Yi-1.5-34B & 01-ai/Yi-1.5-9B & 0.002 \\
01-ai/Yi-1.5-34B & Qwen/Qwen1.5-110B & 0.094 \\
01-ai/Yi-1.5-34B & Qwen/Qwen1.5-32B & 0.061 \\
01-ai/Yi-1.5-34B & Qwen/Qwen1.5-7B & 0.148 \\
01-ai/Yi-1.5-34B & dmis-lab/meerkat-7b-v1.0 & 0.204 \\
01-ai/Yi-1.5-34B & gpt-3.5-turbo-0125 & 0.030 \\
01-ai/Yi-1.5-34B & gpt-4-turbo-2024-04-09 & 0.046 \\
01-ai/Yi-1.5-34B & gpt-4o-2024-05-13 & 0.137 \\
01-ai/Yi-1.5-34B & internistai/base-7b-v0.2 & 0.079 \\
01-ai/Yi-1.5-34B & meta-llama/Meta-Llama-3-70B & 0.026 \\
01-ai/Yi-1.5-34B & meta-llama/Meta-Llama-3-8B & 0.064 \\
01-ai/Yi-1.5-34B & mistralai/Mistral-7B-v0.1 & 0.208 \\
01-ai/Yi-1.5-34B & mistralai/Mixtral-8x7B-v0.1 & 0.075 \\
01-ai/Yi-1.5-9B & Qwen/Qwen1.5-110B & 0.096 \\
01-ai/Yi-1.5-9B & Qwen/Qwen1.5-32B & 0.063 \\
01-ai/Yi-1.5-9B & Qwen/Qwen1.5-7B & 0.146 \\
01-ai/Yi-1.5-9B & dmis-lab/meerkat-7b-v1.0 & 0.202 \\
01-ai/Yi-1.5-9B & gpt-3.5-turbo-0125 & 0.028 \\
01-ai/Yi-1.5-9B & gpt-4-turbo-2024-04-09 & 0.048 \\
01-ai/Yi-1.5-9B & gpt-4o-2024-05-13 & 0.139 \\
01-ai/Yi-1.5-9B & internistai/base-7b-v0.2 & 0.077 \\
01-ai/Yi-1.5-9B & meta-llama/Meta-Llama-3-70B & 0.028 \\
01-ai/Yi-1.5-9B & meta-llama/Meta-Llama-3-8B & 0.062 \\
01-ai/Yi-1.5-9B & mistralai/Mistral-7B-v0.1 & 0.206 \\
01-ai/Yi-1.5-9B & mistralai/Mixtral-8x7B-v0.1 & 0.072 \\
Qwen/Qwen1.5-110B & Qwen/Qwen1.5-32B & 0.033 \\
Qwen/Qwen1.5-110B & Qwen/Qwen1.5-7B & 0.243 \\
Qwen/Qwen1.5-110B & dmis-lab/meerkat-7b-v1.0 & 0.300 \\
Qwen/Qwen1.5-110B & gpt-3.5-turbo-0125 & 0.124 \\
Qwen/Qwen1.5-110B & gpt-4-turbo-2024-04-09 & 0.049 \\
Qwen/Qwen1.5-110B & gpt-4o-2024-05-13 & 0.043 \\
Qwen/Qwen1.5-110B & internistai/base-7b-v0.2 & 0.173 \\
Qwen/Qwen1.5-110B & meta-llama/Meta-Llama-3-70B & 0.068 \\
Qwen/Qwen1.5-110B & meta-llama/Meta-Llama-3-8B & 0.158 \\
Qwen/Qwen1.5-110B & mistralai/Mistral-7B-v0.1 & 0.304 \\
Qwen/Qwen1.5-110B & mistralai/Mixtral-8x7B-v0.1 & 0.169 \\
Qwen/Qwen1.5-32B & Qwen/Qwen1.5-7B & 0.209 \\
Qwen/Qwen1.5-32B & dmis-lab/meerkat-7b-v1.0 & 0.266 \\
Qwen/Qwen1.5-32B & gpt-3.5-turbo-0125 & 0.091 \\
Qwen/Qwen1.5-32B & gpt-4-turbo-2024-04-09 & 0.015 \\
Qwen/Qwen1.5-32B & gpt-4o-2024-05-13 & 0.076 \\
Qwen/Qwen1.5-32B & internistai/base-7b-v0.2 & 0.140 \\
Qwen/Qwen1.5-32B & meta-llama/Meta-Llama-3-70B & 0.035 \\
Qwen/Qwen1.5-32B & meta-llama/Meta-Llama-3-8B & 0.125 \\
Qwen/Qwen1.5-32B & mistralai/Mistral-7B-v0.1 & 0.270 \\
Qwen/Qwen1.5-32B & mistralai/Mixtral-8x7B-v0.1 & 0.136 \\
Qwen/Qwen1.5-7B & dmis-lab/meerkat-7b-v1.0 & 0.056 \\
Qwen/Qwen1.5-7B & gpt-3.5-turbo-0125 & 0.118 \\
Qwen/Qwen1.5-7B & gpt-4-turbo-2024-04-09 & 0.194 \\
Qwen/Qwen1.5-7B & gpt-4o-2024-05-13 & 0.286 \\
Qwen/Qwen1.5-7B & internistai/base-7b-v0.2 & 0.069 \\
Qwen/Qwen1.5-7B & meta-llama/Meta-Llama-3-70B & 0.174 \\
Qwen/Qwen1.5-7B & meta-llama/Meta-Llama-3-8B & 0.084 \\
Qwen/Qwen1.5-7B & mistralai/Mistral-7B-v0.1 & 0.060 \\
Qwen/Qwen1.5-7B & mistralai/Mixtral-8x7B-v0.1 & 0.073 \\
dmis-lab/meerkat-7b-v1.0 & gpt-3.5-turbo-0125 & 0.174 \\
dmis-lab/meerkat-7b-v1.0 & gpt-4-turbo-2024-04-09 & 0.250 \\
dmis-lab/meerkat-7b-v1.0 & gpt-4o-2024-05-13 & 0.343 \\
dmis-lab/meerkat-7b-v1.0 & internistai/base-7b-v0.2 & 0.125 \\
dmis-lab/meerkat-7b-v1.0 & meta-llama/Meta-Llama-3-70B & 0.230 \\
dmis-lab/meerkat-7b-v1.0 & meta-llama/Meta-Llama-3-8B & 0.140 \\
dmis-lab/meerkat-7b-v1.0 & mistralai/Mistral-7B-v0.1 & 0.004 \\
dmis-lab/meerkat-7b-v1.0 & mistralai/Mixtral-8x7B-v0.1 & 0.129 \\
gpt-3.5-turbo-0125 & gpt-4-turbo-2024-04-09 & 0.076 \\
gpt-3.5-turbo-0125 & gpt-4o-2024-05-13 & 0.167 \\
gpt-3.5-turbo-0125 & internistai/base-7b-v0.2 & 0.049 \\
gpt-3.5-turbo-0125 & meta-llama/Meta-Llama-3-70B & 0.056 \\
gpt-3.5-turbo-0125 & meta-llama/Meta-Llama-3-8B & 0.034 \\
gpt-3.5-turbo-0125 & mistralai/Mistral-7B-v0.1 & 0.178 \\
gpt-3.5-turbo-0125 & mistralai/Mixtral-8x7B-v0.1 & 0.045 \\
gpt-4-turbo-2024-04-09 & gpt-4o-2024-05-13 & 0.091 \\
gpt-4-turbo-2024-04-09 & internistai/base-7b-v0.2 & 0.124 \\
gpt-4-turbo-2024-04-09 & meta-llama/Meta-Llama-3-70B & 0.020 \\
gpt-4-turbo-2024-04-09 & meta-llama/Meta-Llama-3-8B & 0.110 \\
gpt-4-turbo-2024-04-09 & mistralai/Mistral-7B-v0.1 & 0.254 \\
gpt-4-turbo-2024-04-09 & mistralai/Mixtral-8x7B-v0.1 & 0.120 \\
gpt-4o-2024-05-13 & internistai/base-7b-v0.2 & 0.216 \\
gpt-4o-2024-05-13 & meta-llama/Meta-Llama-3-70B & 0.111 \\
gpt-4o-2024-05-13 & meta-llama/Meta-Llama-3-8B & 0.201 \\
gpt-4o-2024-05-13 & mistralai/Mistral-7B-v0.1 & 0.348 \\
gpt-4o-2024-05-13 & mistralai/Mixtral-8x7B-v0.1 & 0.212 \\
internistai/base-7b-v0.2 & meta-llama/Meta-Llama-3-70B & 0.105 \\
internistai/base-7b-v0.2 & meta-llama/Meta-Llama-3-8B & 0.015 \\
internistai/base-7b-v0.2 & mistralai/Mistral-7B-v0.1 & 0.129 \\
internistai/base-7b-v0.2 & mistralai/Mixtral-8x7B-v0.1 & 0.004 \\
meta-llama/Meta-Llama-3-70B & meta-llama/Meta-Llama-3-8B & 0.090 \\
meta-llama/Meta-Llama-3-70B & mistralai/Mistral-7B-v0.1 & 0.235 \\
meta-llama/Meta-Llama-3-70B & mistralai/Mixtral-8x7B-v0.1 & 0.101 \\
meta-llama/Meta-Llama-3-8B & mistralai/Mistral-7B-v0.1 & 0.144 \\
meta-llama/Meta-Llama-3-8B & mistralai/Mixtral-8x7B-v0.1 & 0.011 \\
mistralai/Mistral-7B-v0.1 & mistralai/Mixtral-8x7B-v0.1 & 0.133 \\
\bottomrule
\caption{\label{tab:cohend}Measure of effect size between models using Cohen's d on the overall evaluation (English and French included).}
\end{longtable}

\end{document}